\documentclass[conference]{IEEEtran}
\IEEEoverridecommandlockouts

\usepackage{amsmath,amssymb,amsfonts}
\usepackage{mathrsfs}
\usepackage{comment}
\usepackage{graphicx}
\usepackage{cite}
\usepackage[colorlinks=true, linkcolor=blue, citecolor=blue, urlcolor=blue]{hyperref}
\usepackage{color}

\usepackage{multirow}
\usepackage{booktabs}
\usepackage{makecell}
\usepackage{float}
\usepackage{algorithmic}
\usepackage{textcomp}
\usepackage{xcolor}
\makeatletter
\renewcommand{\tagform@}[1]{\maketag@@@{\footnotesize(#1)}}
\makeatother

\def\BibTeX{{\rm B\kern-.05em{\sc i\kern-.025em b}\kern-.08em
    T\kern-.1667em\lower.7ex\hbox{E}\kern-.125emX}}

\begin{document}

\title{Multimodal Doctor-in-the-Loop: A Clinically-Guided Explainable Framework for Predicting Pathological Response in Non-Small Cell Lung Cancer}

\author{
\IEEEauthorblockN{
    Alice Natalina Caragliano\IEEEauthorrefmark{1},
    Claudia Tacconi\IEEEauthorrefmark{2},
    Carlo Greco\IEEEauthorrefmark{2}\IEEEauthorrefmark{3},
    Lorenzo Nibid\IEEEauthorrefmark{2}\IEEEauthorrefmark{4},
    Edy Ippolito\IEEEauthorrefmark{2}\IEEEauthorrefmark{3},\\
    Michele Fiore\IEEEauthorrefmark{2}\IEEEauthorrefmark{3},
    Giuseppe Perrone\IEEEauthorrefmark{2}\IEEEauthorrefmark{4},
    Sara Ramella\IEEEauthorrefmark{2}\IEEEauthorrefmark{3},
    Paolo Soda\IEEEauthorrefmark{1}\IEEEauthorrefmark{5}, and
    Valerio Guarrasi\IEEEauthorrefmark{1}
}
\\
\IEEEauthorblockA{\IEEEauthorrefmark{1}\textit{Research Unit of Computer Systems and Bioinformatics, Department of Engineering,} \\
    \textit{Università Campus Bio-Medico di Roma, Rome, Italy} \\ \{a.caragliano, valerio.guarrasi, p.soda\}@unicampus.it}
   
\IEEEauthorblockA{\IEEEauthorrefmark{2} \textit{Fondazione Policlinico Universitario Campus Bio-Medico, Rome, Italy}} 

\IEEEauthorblockA{\IEEEauthorrefmark{3}\textit{Research Unit of Radiation Oncology, Department of Medicine and Surgery,} \\
    \textit{Università Campus Bio-Medico di Roma, Rome, Italy}}
\IEEEauthorblockA{\IEEEauthorrefmark{4}\textit{Research Unit of Anatomical Pathology, Department of Medicine and Surgery,} \\ \textit{Università Campus Bio-Medico di Roma, Rome, Italy}}
\IEEEauthorblockA{\IEEEauthorrefmark{5}\textit{Department of Diagnostics and Intervention, Radiation Physics, Biomedical Engineering,} \\ \textit{Umeå University, Umeå, Sweden}}
}

\maketitle

\begin{abstract}
This study proposes a novel approach combining Multimodal Deep Learning with intrinsic eXplainable Artificial Intelligence techniques to predict pathological response in non-small cell lung cancer patients undergoing neoadjuvant therapy. Due to the limitations of existing radiomics and unimodal deep learning approaches, we introduce an intermediate fusion strategy that integrates imaging and clinical data, enabling efficient interaction between data modalities. The proposed \textit{Multimodal Doctor-in-the-Loop} method further enhances clinical relevance by embedding clinicians' domain knowledge directly into the training process, guiding the model's focus gradually from broader lung regions to specific lesions. Results demonstrate improved predictive accuracy and explainability, providing insights into optimal data integration strategies for clinical applications.
\end{abstract}


\section{Introduction} \label{sec:introduction}

Non-small cell lung cancer (NSCLC) is the most common subtype of lung cancer, constituting approximately 85\% of lung cancer cases~\cite{cancernet2022}. Currently, surgery remains the main treatment for early-stage and resectable locally advanced NSCLC, although a notable number of patients experience post-surgery recurrence. Neoadjuvant therapy (NAT) has shown potential in improving overall survival rates and reducing the risk of distant disease recurrence ~\cite{betticher2006prognostic}. Achieving a complete pathological response after NAT, indicating the absence of tumor cells in all specimens, may have a potential prognostic role and serve as a surrogate survival endpoint ~\cite{rosner2022association}. 

Evaluating pathological response before surgical resection provides valuable insights into tumor sensitivity to the administered therapy, enabling clinicians to tailor the type of treatment to the needs of patients and reserve surgical interventions only for patients who are most likely to benefit from them. 
However, achieving a complete pathological response is relatively rare in NSCLC patients~\cite{hellmann2014pathological}. 
In contrast, major pathological response, defined as the presence of no more than 10\% viable tumor cells, is observed in a larger proportion of NSCLC patients and has been associated with significant clinical benefits, including improved progression-free survival~\cite{provencio2020neoadjuvant}. 
Additionally, its  higher prevalence enables the identification of a broader cohort of patients who may benefit from NAT~\cite{hellmann2014pathological}. Notably, considering a slightly higher threshold for viable tumor cells also addresses potential inter- and intra-observer variability in pathological evaluations~\cite{weissferdt2020agreement}. 
Consequently, major pathological response represents a more achievable, reliable, and clinically relevant endpoint for assessing NAT efficacy~\cite{hellmann2014pathological}. 
Therefore, here we selected major pathological response as the predicting endpoint, using the term pathological response (pR) to refer to both complete and major pathological response.

State-of-the-art reports that radiomics features extracted from Computed Tomography (CT) scans~\cite{coroller2017radiomic,liu2023development} correlate with pR. However, this approach has limitations: while non-invasive, it relies on hand-crafted features, which may not sufficiently capture tumor complexity and variability~\cite{sun2024pet}.

A powerful solution to address these issues is to leverage Deep Learning (DL), which has demonstrated notable success in cancer-related tasks, such as tumor segmentation, diagnosis, and classification~\cite{cellina2022artificial}. Recent works have used DL to predict pR~\cite{qu2024non, ye2024non}, relying on a unimodal approach based only on imaging data. However, this approach may fail to fully exploit complementary information from different data sources. In contrast, Multimodal Deep learning (MDL), which integrates various types of data, such as imaging and clinical data, has the potential to improve predictive performance by capturing a more comprehensive representation of tumor features~\cite{sangeetha2024enhanced}.

Among MDL strategies, fusion techniques play a crucial role in determining how different modalities interact: \textit{early fusion} combines data from multiple modalities at the data level; \textit{intermediate fusion}~\cite{guarrasi2025systematic} merges modality-specific features at one or more points within the model, allowing joint representation learning~\cite{francesconi2025class, ruffini2024multi, di2025graph, guarrasi2023multi}; \textit{late fusion} integrates outputs from modality-specific models at the decision level~\cite{mogensen2025optimized, guarrasi2022optimized, caruso2022multimodal}. Prior studies on pR have explored MDL via \textit{early fusion}\cite{lin2022ct} and \textit{late fusion}\cite{she2022deep}. However, these approaches may fail to fully exploit feature interactions. To address this, we propose an \textit{intermediate fusion} approach, enabling interactions between imaging and clinical features during network training, enhancing efficient multimodal information integration.

Despite MDL's potential, the explainability of deep model results remains a major challenge, especially in healthcare~\cite{hou2024self, guarrasi2024multimodal}, where understanding \textit{what} the model focuses on is as important as the prediction itself. While some studies have applied eXplainable Artificial Intelligence (XAI) techniques to analyze pR predictions~\cite{she2022deep, ye2024non}, they relied on \textit{post-hoc} methods applied after training, limiting faithfulness to the model's actual decision-making process. In contrast, \textit{intrinsic} methods embed explainability directly into the model architecture during training, providing transparent explanations of the model’s outcomes~\cite{hou2024self, caragliano2025doctor}. To the best of our knowledge, no prior work in the context of pR has explored combining MDL via intermediate fusion with \textit{intrinsic} explainability.

Moreover, the limited size of medical datasets complicates the training of DL models that generally benefits from large amounts of data. Hence, in small datasets and complex tasks like pR prediction, guiding the model towards clinically relevant areas is crucial, as it may struggle to identify them independently. To address this, we propose the \textit{Multimodal Doctor-in-the-Loop} method that leverages XAI techniques to integrate domain knowledge into the training process, guiding the model's focus toward clinically relevant regions. Additionally, our approach progressively refines the model's focus from a broader lung region to the lesion itself. This \textit{multi-view} approach distinguishes our work from state-of-the-art studies, typically analyzing only a single view (the lesion region)~\cite{coroller2017radiomic, qu2024non,  ye2024non, lin2022ct, she2022deep}.
The main contributions of this work are:
\begin{itemize}
\item We propose a technique leveraging MDL and \textit{intrinsic} XAI to achieve non-invasive, robust, and explainable pR predictions in NSCLC patients undergoing NAT.
\item We introduce the \textit{Multimodal Doctor-in-the-Loop}, a \textit{gradual multi-view} framework that incorporates  domain knowledge to guide the model's focus from broad to specific clinically relevant regions, enabling comprehensive analysis and addressing task complexity.
\item We compare intermediate fusion with early and late fusion strategies, providing insights into optimal integration of clinical and imaging data for pR prediction.
\end{itemize}

\section{Methods} \label{sec:methods}

Traditional neural network training paradigms often optimize for specific loss functions to improve accuracy, without providing insights into their decision-making processes, potentially leading to trust issues, especially in critical fields like healthcare. While models trained on large datasets can autonomously learn which features to focus on to provide accurate predictions, in medical contexts, where annotated datasets are typically small and difficult to collect, the network may struggle to identify the most relevant features independently. This challenge is compounded by the complexity of medical tasks, where multiple data modalities are critical for accurate diagnosis. Thus, it is crucial to guide the model's focus through domain-specific insights, ensuring that the model focuses on clinically relevant areas identified by domain experts, while also leveraging the enriched feature representation provided by MDL. Indeed, relying solely on imaging data may limit the model’s ability to capture a comprehensive representation of a patient’s condition, as the model may overlook valuable clinical context. By effectively integrating clinical data, which often contain critical complementary information, the model can develop a more holistic understanding of disease patterns. 

\begin{figure}[h] 
    \centering 
    \includegraphics[width=0.99\columnwidth]{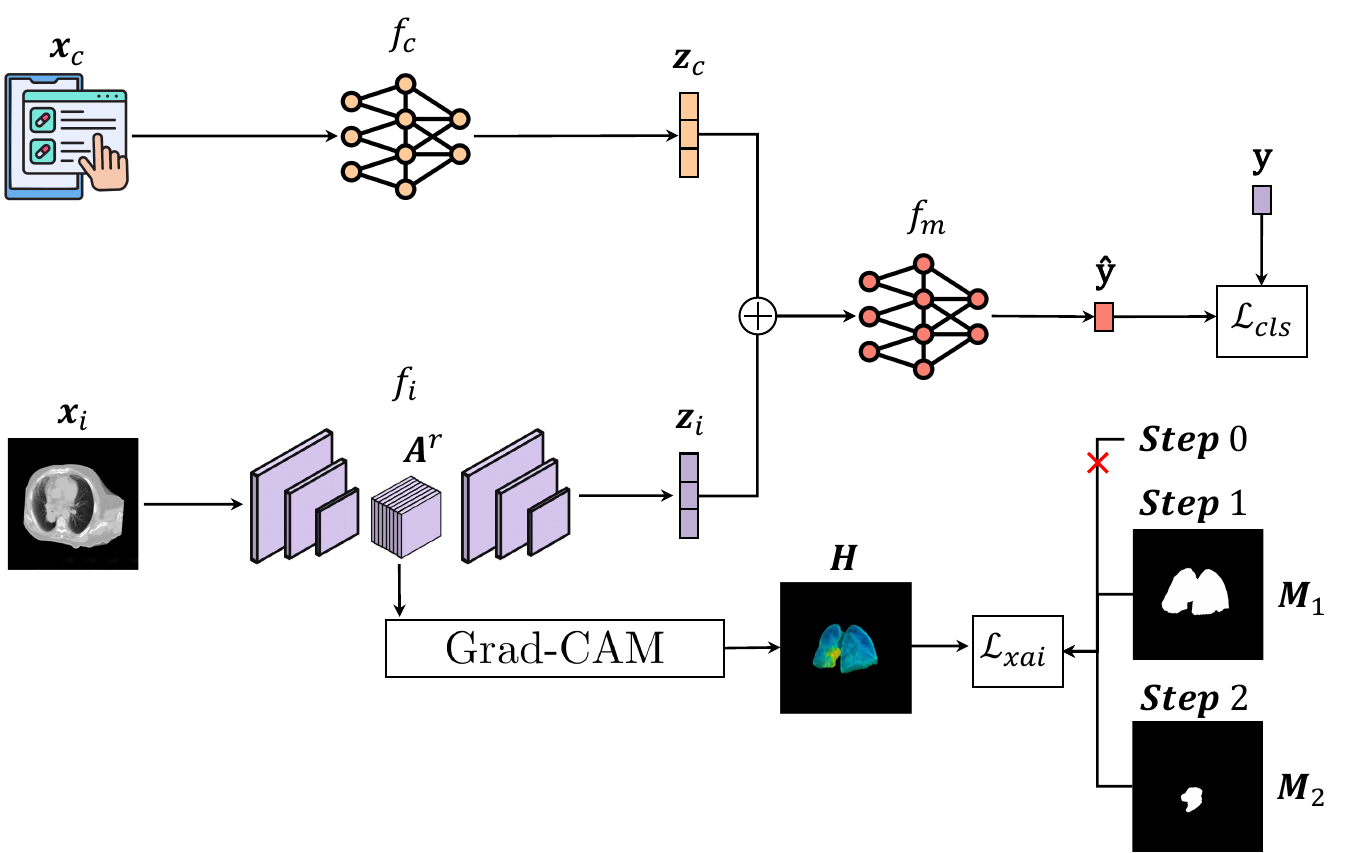} 
    \caption{{\textit{Multimodal Doctor-in-the-Loop} training paradigm: integration of clinical and imaging data. \textit{Step 0}: Model trained on global image using classification loss ($L_{cls}$). \textit{Step 1}: Segmentation mask $\boldsymbol{M}_{1}$ guides model focus via Grad-CAM heatmaps, training with classification ($L{cls}$) and XAI ($L_{xai}$) losses. \textit{Step 2}: Detailed segmentation mask $\boldsymbol{M}_{2}$ further refines focus using classification ($L{cls}$) and XAI ($L_{xai}$) losses.}}
    \label{fig:method} 
\end{figure}

As shown in \hyperref[fig:method]{Fig.~\ref{fig:method}}, to address these challenges, we introduce the \textit{Multimodal Doctor-in-the-Loop} training paradigm, which extends the unimodal \textit{Doctor-in-the-Loop} methodology~\cite{caragliano2025doctor} through intermediate fusion of imaging and clinical data. The unimodal \textit{Doctor-in-the-Loop} approach is trained exclusively on CT images using a Gradual Learning (GL) process, which progressively integrates multiple CT views. These views range from broad anatomical structures to more detailed areas defined by expert-provided segmentation masks, effectively incorporating domain knowledge throughout the training process. Separately, a clinical model is trained on patient-specific clinical data. The intermediate fusion strategy is then employed to combine these unimodal models into a joint multimodal architecture, enabling the fusion of imaging and clinical information. This fusion leads to a more comprehensive representation of tumor characteristics, enhancing predictive robustness while maintaining transparency.

The proposed paradigm represents a shift from traditional training methods by adapting to the task complexity over time, while continuously guiding the model's learning trajectory with expert insights. This approach not only enhances the model’s accuracy but also boosts explainability and adaptability in real-world applications where expert knowledge and multimodal data play a crucial role.

The \textit{Multimodal Doctor-in-the-Loop} framework integrates CT images and clinical data through an intermediate fusion strategy, where modality-specific features from unimodal models are combined.
Let $\boldsymbol{X}_i \subseteq \mathbb{R}^{H \times W \times D}$ denote the imaging input space, where each image $\boldsymbol{x}_i \in \boldsymbol{X}_i$ has dimensions $H$ (height), $W$ (width), and $D$ (depth), with $D$ representing the number of slices in $\boldsymbol{x}_i$. Imaging features are extracted via a fully convolutional module $f_i$.
Similarly, let $\boldsymbol{X}_c \subseteq \mathbb{R}^{F}$ denote the clinical input feature space, where each clinical feature vector $\boldsymbol{x}_c \in \boldsymbol{X}_c$ comprises $F$ clinical variables. Clinical features are extracted through the multilayer perceptron (MLP)-based module $f_c$.
The multimodal fusion module $f_m$ receives the concatenated imaging and clinical features as a single combined feature vector and maps it, via fully connected (FC) layers, to the label space $\boldsymbol{Y}$, represented by one-hot encoded vectors corresponding to the classes ${1, 2, \dots, L}$.

\subsection{CT Model}
\paragraph{Model Formulation}
The CT model is trained by using the unimodal \textit{Doctor-in-the-Loop} approach on CT scans. The convolutional neural network is characterized by its parameters $\boldsymbol{\theta}_{i}$, and is designed to perform image classification. The network consists of the module $f_i$, followed by a FC layer that produces a vector of class scores, also known as logits. A softmax function is then applied to these logits to produce the predicted class probabilities $\boldsymbol{\hat{y}}$.

Alongside the primary classification task, two expert-provided segmentation masks ($\boldsymbol{M}_1, \boldsymbol{M}_2,$) are employed to guide the learning process. Each mask $\boldsymbol{M}_t$ is represented as $\boldsymbol{M}_t \subseteq \mathbb{R}^{H \times W \times D}$, and corresponds to a different view with $t$ indexing levels of granularity from macroscopic ($\boldsymbol{M}_1$) to microscopic ($\boldsymbol{M}_2$) focus areas. These masks highlight regions of interest in the images, informing the network about regions particularly relevant for the classification task. In this study, we used two segmentation masks, corresponding to the lung and lesion areas. It is important to note that these masks are used exclusively in the training phase to guide the learning process and are not required in the testing phase. During this \textit{gradual multi-view} training phase, the network first aims to achieve high accuracy in image classification, learning from global image content, corresponding to a bounding box around the lung region. The network then leverages segmentation masks to iteratively refine its focus towards the areas of the image identified as critical by the experts. Specifically, as training progresses, the integration of the Grad-CAM heatmaps, which highlight the regions of the image most influential for the network's predictions, with the expert masks ensures the model aligns its focus with the expert-identified regions. 

\paragraph{Loss Functions}
To train our model effectively, we employ a composite loss function that target both prediction accuracy and explainability. In detail, we use a classification loss $\mathscr{L}_{\mathit{cls}}$, whose primary objective is to correctly classify the input images based on the learned features, and an XAI loss $\mathscr{L}_{\mathit{xai}}$, which encourages the model to align its focus with expert-provided regions of interest. Hence, during training, the network optimizes the composite loss function $\mathscr{L}$, defined as:
\begin{equation}
\resizebox{0.92\columnwidth}{!}{$
    \mathscr{L}(\boldsymbol{\theta}_k; \boldsymbol{x}_k, \boldsymbol{y}, \boldsymbol{M}_t) = \mathscr{L}_{\mathit{cls}}(\boldsymbol{\theta}_k; \boldsymbol{x}_k, \boldsymbol{y}) + \lambda \mathscr{L}_{\mathit{xai}}(\boldsymbol{\theta}_k; \boldsymbol{x}_i, \boldsymbol{M}_t)\label{L}
    $}
\end{equation}
where $k=i$, $\lambda$ is a hyperparameter that balances the contribution of the XAI in the overall training objective, and $\boldsymbol{y}$ is the one-hot encoded class vector. 
Initially, the model is trained using only $\mathscr{L}_{cls}$ ($\lambda = 0$). Once the model reaches a convergence criterion based on its validation performance, $\mathscr{L}_{xai}$ is introduced ($\lambda = 1$). The composite loss function $\mathscr{L}$ balances both classification performance and explainability, ensuring the network not only provides accurate predictions but also offers insights into its decision-making process. 

\textit{Classification Loss}: The classification loss $\mathscr{L}_{\mathit{cls}}$ is a fundamental component used for training the network to correctly predict the class labels of input images. $\mathscr{L}_{\mathit{cls}}$, computed using the cross-entropy loss function, measures the discrepancy between the predicted probabilities and the actual class labels:
\begin{equation}
    \mathscr{L}_{\mathit{cls}}(\boldsymbol{\theta}_{k}; \boldsymbol{x}_k, \boldsymbol{y}) = - \sum_{l=1}^L y_l \log(\hat{y}_l)\label{Lcls}
\end{equation}
where $k=i$, $\boldsymbol{\theta}_{i}$, $\boldsymbol{x}_i$ and $\boldsymbol{y}$ are the network parameters, the input image and the class vector already defined, and ${y}_l$ is the indicator for class $l$.
The vector $\boldsymbol{\hat{y}}$ corresponds to the predicted class probabilities obtained via the softmax function, with $\hat{y}_l$ representing the posterior probability for class $l$.

\textit{XAI loss}: The XAI loss $\mathscr{L}_{\mathit{xai}}$ enhances the model's transparency by encouraging alignment between the network-generated heatmaps and the expert-provided masks. It utilizes the mean squared error (MSE) to quantify the difference between the Grad-CAM heatmap $\boldsymbol{H}(\boldsymbol{x}_i;\boldsymbol{\theta}_{i})$ and the expert mask $\boldsymbol{M}_t$.
The heatmap $\boldsymbol{H}(\boldsymbol{x}_i;\boldsymbol{\theta}_{i})$, generated based on the gradients of the target class with respect to each feature map $\boldsymbol{A}^r$ of a target layer of the network, highlights important pixels influencing the network’s prediction.
$\mathscr{L}_{\mathit{xai}}$ is defined as:
\begin{equation}
    \mathscr{L}_{\mathit{xai}}(\boldsymbol{\theta}_{k}; \boldsymbol{x}_i, \boldsymbol{M}_t) = \frac{1}{N} \sum_{n=1}^N (M_{t,n} - H_n(\boldsymbol{x}_i; \boldsymbol{\theta}_{k}))^2\label{Lxai}
\end{equation}
where $k=i$, $N$ is the total number of pixels in the image, $M_{t,n}$ is the value at pixel $n$ in the expert mask $\boldsymbol{M}_t$, and $H_n(\boldsymbol{x}_i; \boldsymbol{\theta}_{i})$ represents the value at pixel $n$ in the heatmap. It is important to highlight that $\mathscr{L}_{\mathit{xai}}$ always depends on $\boldsymbol{x}_i$, as it is computed over imaging-based Grad-CAM heatmaps.

Grad-CAM~\cite{bib:selvaraju2017grad} is used to generate heatmaps $\boldsymbol{H}(\boldsymbol{x}_i;\boldsymbol{\theta}_{i})$ that visually indicate which parts of the input image are most important for the predictions made by the neural network. The first convolutional layer was selected as target layer for the computation of feature maps, due to its ability to highlight low-level features, such as edges and textures. In this way, we focus on features that are more directly comparable to the expert-provided segmentation masks. In contrast, using a deeper layer would capture more abstract features that may be less relevant for  segmentation masks comparison. 

\subsection{Clinical Model}
\paragraph{Model Formulation}
The clinical model is formulated as an MLP with trainable parameters $\boldsymbol{\theta}_{c}$. The network consists of the $f_c$ module, followed by a FC layer and a softmax function to produce the predicted class probabilities.  

\paragraph{Loss Function}
The clinical model is trained using the cross-entropy loss function $\mathscr{L}_{\mathit{cls}}$, as defined in \eqref{Lcls}, with $k=c$.

\subsection{Multimodal Doctor-in-the-Loop Model}
\paragraph{Model Formulation}
Since imaging and clinical features provide complementary information, once the unimodal models (CT and clinical) have been trained separately, we integrate them through an intermediate fusion strategy. 

Let $\boldsymbol{x}_m = (\boldsymbol{x}_i, \boldsymbol{x}_c)$ denote the combined multimodal input, consisting of the imaging and clinical data, $\boldsymbol{z}_i = f_i(\boldsymbol{x}_i; \boldsymbol{\theta}_i)$ represent the feature vector extracted from the CT model, $\boldsymbol{z}_c = f_c(\boldsymbol{x}_c; \boldsymbol{\theta}_c)$ denote the feature vector obtained from the clinical model, and $\boldsymbol{\theta}_m$ denote the trainable parameters of the multimodal model. The intermediate fusion process involves the $f_m$ module concatenating the feature vectors $\boldsymbol{z}_i$ and $\boldsymbol{z}_c$, and passing the fused feature vector through an MLP with a final FC layer, followed by a softmax function, which produces the predicted class probabilities $\boldsymbol{\hat{y}}$. Crucially, the multimodal model is trained end-to-end, enabling the joint optimization of modality-specific representations and their integration.  

\paragraph{Loss Function}
The multimodal model is trained using the weighted composite loss function $\mathscr{L}$, defined in \eqref{L}, with  $k=m$. $\mathscr{L}_{\mathit{xai}}$ ensures explainability by aligning the Grad-CAM heatmaps with the same expert-provided masks used in the unimodal CT model.

\paragraph{Training Procedure}
The training of the \textit{Multimodal Doctor-in-the-Loop} model follows a two-stage process: 
\begin{enumerate}

\item \textit{Unimodal Training}: Both unimodal models (CT and Clinical) are initially trained separately on their respective data sources. This phase focuses on learning relevant features from each modality independently.
\item \textit{Intermediate Fusion and Joint Training}: After the unimodal models are trained, they are fused into the multimodal model, which is trained using both imaging and clinical data. The training uses the combined $\mathscr{L}$ function, ensuring that the model not only learns to make accurate predictions but also aligns its focus with expert-provided regions of interest.
\end{enumerate}

\section{Experimental Setup} \label{sec:experiments}

To validate the proposed \textit{Multimodal Doctor-in-the-Loop} approach, we conducted a series of experiments using an in-house dataset of NSCLC patients for the pR prediction task. 

\subsection{Dataset} The experiments were performed using a dataset of 100 NSCLC patients with TNM stage II-III~\cite{edition2017ajcc}, collected at \textit{Fondazione Policlinico Universitario Campus Bio-Medico of Rome}. All patients underwent neoadjuvant chemoradiation therapy followed by surgical resection. Among these patients, 36\% achieved a pR, defined as having no more than 10\% viable tumor cells in all specimens. 

This study was approved by two separate Ethical Committees. The retrospective phase was approved on October 30, 2012, and registered on ClinicalTrials.gov on July 12, 2018 (identifier: NCT03583723). The prospective phase was approved with the identifier 16/19 OSS. All patients provided written informed consent. 
Data used in this study are available from the corresponding author upon reasonable request.

\paragraph{Imaging data}
For each patient, chest CT scans acquired within one month before the start of NAT were collected. These CT scans were annotated by expert radiation oncologists, as part of the treatment planning process. Specifically, the clinicians provided segmentation masks for both lung and lesion regions, and we used the Planning Tumor Volume as the lesion segmentation. 

\paragraph{Clinical data}
A comprehensive set of clinical features was collected for each patient to capture their overall health status and treatment parameters. These features encompass: (i) \textit{general patient characteristics}, including sex, age, weight, height, smoking status, number of cigarettes per day, family history of tumors, numeric pain rating scale (NRS), and comorbidities; (ii) \textit{tumor information}, including diagnosis, tumor stage, and TNM classification; (iii) \textit{biopsy details}, including diagnosis material, sampling site and technique; (iv) \textit{molecular biomarkers}, including EGFR, ALK, and PD-L1; (v) \textit{treatment details}, including induction chemotherapy administration and scheme, number of induction chemotherapy cycles, chemotherapy scheme,  total and daily delivered radiation dose, number of radiotherapy sessions, radiotherapy technique and duration, days of treatment interruption, occurrence of permanent treatment interruption, and toxicities (esophageal, pulmonary, and hematologic).

\subsection{Pre-processing} 

\paragraph{Imaging data}
To standardize the data and optimize model performance, a series of pre-processing steps were applied. First, all CT images were resampled to a uniform resolution of 1 $\times$ 1 $\times$ 1~$mm^{3}$ via a {nearest-neighbor} interpolation, ensuring consistency across the dataset. Images were then clipped, using a lung window setting (mean: -300~HU; width: 1200~HU) to minimize the impact of extreme values, and linearly normalized to standardize the pixel intensity distribution, scaling the values to the [0, 1] range. Finally, CT scans were cropped, using a fixed rectangular bounding-box encompassing the lung region, to ensure inputs of uniform size (324 $\times$ 324 pixels). To enhance model robustness and mitigate overfitting, data augmentation was applied using spatial {shifts} (±3 pixels) and vertical {flips}, without altering the underlying anatomical structures.

\paragraph{Clinical data}
To ensure consistency during model training, all clinical feautures underwent a standardized pre-processing pipeline: categorical features (e.g., comorbidities) were one-hot encoded to convert them into binary vectors; ordinal features (e.g., disease stage) were encoded based on their inherent ranking; numerical features (e.g., patient age) were standardized using z-score normalization. 

\subsection{Experimental Configuration} \label{sec:experiments_configuration}
To assess the effectiveness of our method, we conducted a series of experiments, including alternative fusion strategies to evaluate the impact of intermediate fusion compared to early and late fusion. Additional experiments were designed to analyze the contributions of GL and XAI to the performance of the proposed model. A detailed description of each experiment is provided below. 

\paragraph{Unimodal CT}
This experiment evaluates the unimodal \textit{Doctor-in-the-Loop} approach, trained on CT images. The CT model follows a three-stage progressive refinement, integrating both GL and XAI guidance: \textit{Step 0}, where the model is trained on the global image, using only $\mathscr{L}_{cls}$; \textit{Step 1}, where the training is refined by introducing the lung segmentation mask as a guide for the model's focus, applying $\mathscr{L}$; \textit{Step 2}, where the model’s focus is further refined by using the lesion segmentation as a guidance, again applying $\mathscr{L}$.

\paragraph{Unimodal Clinical}
This experiment evaluates the unimodal clinical model, trained on clinical features, applying only $\mathscr{L}_{cls}$.  It serves as a baseline for the multimodal model and provides insight into the predictive value of patient-specific variables when used independently.
\paragraph{Multimodal Doctor-in-the-Loop} This experimental configuration evaluates the intermediate fusion strategy, in which modality-specific representations from both the clinical model and \textit{Step 2} of the unimodal CT model are merged. This fusion strategy enables joint representation learning while preserving the individual processing strengths of each modality. As in the unimodal \textit{Doctor-in-the-Loop}, the combined loss $\mathscr{L}$ is applied, incorporating XAI to guide the model's focus. 

\paragraph{Early fusion}
This experiment evaluates an alternative configuration of the \textit{Multimodal Doctor-in-the-Loop}, where CT images and clinical features are combined at the data level via early fusion. The fused data are jointly processed, enabling feature interactions from both modalities at the earliest stage of learning, with the multimodal model trained using $\mathscr{L}_{cls}$. 

\paragraph{Late fusion}
This experiment evaluates another alternative configuration of the \textit{Multimodal Doctor-in-the-Loop}, in which the CT and clinical models are trained independently, and their outputs are integrated at the decision level via late fusion. The final prediction is obtained by combining the probability scores from both models, leveraging each modality’s independent contribution to the classification task.

\paragraph {XAI-guide}
This ablation study isolates the impact of XAI guidance within the \textit{Doctor-in-the-Loop} framework by removing the GL component. Specifically, the CT model is trained directly using the lesion segmentation mask as a fixed guide throughout the process, applying the loss function $\mathscr{L}$. Consequently, the model does not adjust its learning trajectory over time through the progression of different stages of focus. We then apply early, intermediate, and late fusion using this XAI-guided CT model alongside the clinical model.  

\paragraph{\textit{{Segmentation}}}

In this ablation study, the CT model is trained directly on the expert-provided lesion segmentation mask, bypassing both XAI and GL, and applying only $\mathscr{L}_{cls}$. As in the \textit{XAI-guide}, we apply early, intermediate, and late fusion using this CT model alongside the clinical model. 

\subsection{Training}
For all CT-based models, including the unimodal CT, ablation studies, and the multimodal configuration, we used a \textit{DenseNet169}, a 3D convolutional neural network, as it demonstrated successful results in lung cancer-related tasks~\cite{zhu2020ct}. An MLP architecture was employed for the clinical model, for the early fusion, and as the fusion module in the intermediate fusion setup. The dataset was split into training (60\%), validation (20\%) and test (20\%) sets, via a 5-fold stratified cross-validation scheme. All experiments used consistent data splits and training configurations, as described below. 

Optimization was carried out using the Adam optimizer, with an initial learning rate of 0.001, and a weight decay of 0.00001. Training included a 50-epoch warm-up period and was limited to a maximum of 300 epochs, with early stopping applied if validation loss failed to improve for 50 consecutive epochs, thereby preventing overfitting. When employed, the hyperparameter $\lambda$ was empirically determined to balance classification performance and explainability. A range of values was explored (from 0.1 to 2, with a step size of 0.1), and the final choice of $\lambda = 1$ was selected for optimal performance.

The source code for the proposed \textit{Multimodal Doctor-in-the-Loop} framework is available in the project's GitHub repository at \url{https://github.com/cosbidev/Doctor-in-the-Loop}.
\subsection{Evaluation Metrics} To comprehensively evaluate the models' performance, we employed the following metrics: Accuracy (ACC), Area Under the Curve (AUC) and Matthew’s Correlation Coefficient (MCC). Results are reported as the mean and standard error, computed across the different folds. 

\section{Results and Discussion}
In this section, we present and discuss the results of all conducted experiments. Quantitative performance metrics are detailed in \hyperref[table:performance_results] {Table~\ref{table:performance_results}}, which reports ACC, AUC, and MCC, for all the experiments. For each metric, the best performing value is highlighted in bold. Visual comparisons of performance metrics between the \textit{Multimodal Doctor-in-the-Loop} approach and each experiment are provided in \hyperref[fig:plot1]{Fig.~\ref{fig:plot1}}. 
 
\begin{table}[h]
\caption{Performance metrics (ACC, AUC, and MCC) under the different experimental configurations. For each metric, the best performing value is highlighted in bold.}
\label{table:performance_results}
\centering
\renewcommand{\arraystretch}{1.25}
\resizebox{\columnwidth}{!}{
\begin{tabular}{lcllll}
\toprule
\textbf{Experiment} & \textbf{Fusion strategy} & \textbf{ACC (\%)} & \textbf{AUC (\%)} & \textbf{MCC (\%)} \\ \midrule
\multirow{1}{*}{\textit{Unimodal CT}} & $-$  & 62.00\text{\tiny ±1.22} & 58.35\text{\tiny ±4.18} & 16.61\text{\tiny ±3.29} \\
                                    \midrule
\multirow{1}{*}{\textit{Unimodal Clinical}} & $-$  & 67.00\text{\tiny ±2.55} & 70.19\text{\tiny ±5.27} & 28.04\text{\tiny ±6.10} \\
                                    \midrule
\multirow{3}{*}{\shortstack[l]{\textit{Multimodal}\\\textit{Doctor-in-the-Loop}}} & \textit{Early}  & 69.00\text{\tiny ±3.32} & 71.08\text{\tiny ±4.72} & 31.92\text{\tiny ±6.83} \\
                                    & \textit{Intermediate} &  \textbf{73.00}\text{\tiny ±2.00} & \textbf{72.49}\text{\tiny ±3.21} & \textbf{40.30}\text{\tiny ±4.51} \\
                          & \textit{Late} &  {65.00}\text{\tiny ±3.53} & {70.26}\text{\tiny ±2.33} & {23.72}\text{\tiny ±6.91}  \\ \midrule

\multirow{3}{*}{\textit{XAI-guide}} & \textit{Early}  & 68.00\text{\tiny ±5.14} & 67.19\text{\tiny ±4.11} & 31.47\text{\tiny ±9.36} \\
                                    & \textit{Intermediate} &  69.00\text{\tiny ±1.87} & 71.05\text{\tiny ±2.68} & 28.11\text{\tiny ±7.74} \\
                          & \textit{Late} &  {62.00}\text{\tiny ±3.00} & {68.23}\text{\tiny ±3.05} & {18.63}\text{\tiny ±6.04}  \\ \midrule

\multirow{3}{*}{\textit{Segmentation}} & \textit{Early}  & 58.00\text{\tiny ±6.44} & 56.39\text{\tiny ±3.73} & 12.84\text{\tiny ±11.27} \\
                                    & \textit{Intermediate} &  62.00\text{\tiny ±2.00} & 58.57\text{\tiny ±6.65} & 12.84\text{\tiny ±6.32} \\
                          & \textit{Late} &  {45.00}\text{\tiny ±5.24} & {46.90}\text{\tiny ±5.25} & {-23.27}\text{\tiny ±7.42}  \\ \bottomrule

\end{tabular}
}
\end{table}
\subsection{\textit{Multimodal vs Unimodal}}
As shown in \hyperref[table:performance_results]{Table~\ref{table:performance_results}}, the proposed \textit{Multimodal Doctor-in-the-Loop} approach, based on the intermediate fusion strategy, achieves the best overall performance across all experimental configurations. These results reflect the combined impact of intermediate multimodal data fusion, along with GL strategy and XAI guidance, which are intrinsic to the \textit{Doctor-in-the-Loop} methodology. The proposed \textit{Multimodal Doctor-in-the-Loop} substantially outperforms both the CT model and the clinical model. The substantial increase in MCC highlights the improved ability of the proposed model to distinguish between patients who achieve pR from those who do not respond well to therapy. The top row of \hyperref[fig:plot1]{Fig.~\ref{fig:plot1}} highlights the performance gains achieved by integrating clinical and imaging modalities within the \textit{Doctor-in-the-Loop} framework.

\begin{figure*}[!t] 
    \centering
    \includegraphics[width=1.46\columnwidth]{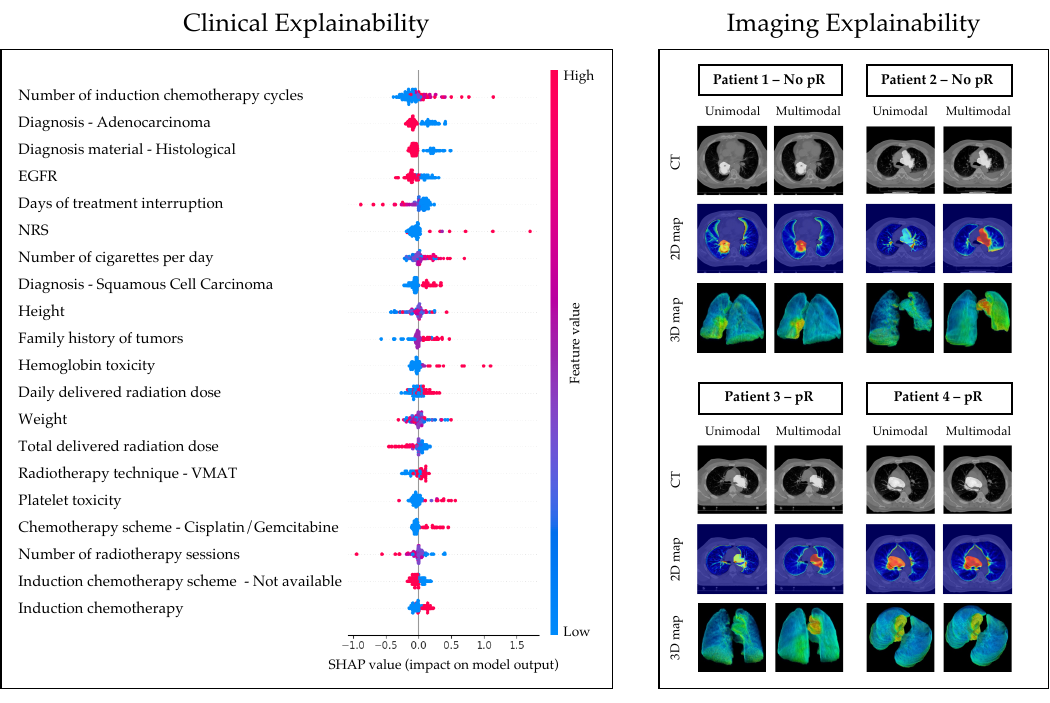} 
    \caption{{\textit{Clinical Explainability}: SHAP values of the 20 most significant clinical features, highlighting variables related to diagnosis, patient characteristics, and treatments. A color gradient is used to indicate the relevance of each feature (low: blue, high: red). \textit{Imaging Explainability}: Grad-CAM heatmaps comparing unimodal and multimodal \textit{Doctor-in-the-Loop} approaches for four representative patients categorized as \textit{No-pR} or \textit{pR}. For each patient, three visualizations are shown: the CT image with lesion segmentation, the corresponding 2D Grad-CAM map, and the corresponding 3D Grad-CAM map, with a color gradient indicating the heatmap intensity values (low: blue, high: red). The multimodal model demonstrates enhanced explainability through more refined focus.}}
    \label{fig:expl}    
\end{figure*}

The effectiveness of the \textit{Multimodal Doctor-in-the-Loop} framework is further supported by its improved explainability compared to the unimodal CT model (\hyperref[fig:expl] {Fig.~\ref{fig:expl}}). To evaluate the model's focus, post-hoc Grad-CAM heatmaps were generated for both unimodal and multimodal models. Since both are based on the \textit{Doctor-in-the-Loop} methodology, they incorporate expert segmentation guidance which refines the model’s focus on the lesion region. While the heatmaps of both models highlight the lesion, those of the multimodal model exhibit higher intensity values and a more refined focus in the lesion area, ensuring more precise localization of the lesion itself. This demonstrates an improvement in explainability compared to the unimodal CT model. Additionally, the 3D heatmaps, thanks to their comprehensive view, provide further insights, revealing that the multimodal model maintains a more precise and consistent focus across all slices, regardless of the presence or absence of pR. It is also worth noting that some cases (e.g., Patient 4) exhibit similar explainability patterns across both models. 
The improvement in the Grad-CAM heatmaps (\hyperref[fig:expl] {Fig.~\ref{fig:expl}}) aligns with the quantitative performance gains observed for the multimodal model in comparison to the unimodal one (\hyperref[table:performance_results] {Table~\ref{table:performance_results}}, \hyperref[fig:plot1] {Fig.~\ref{fig:plot1}}).

In addition to imaging-based explainability, \hyperref[fig:expl]{Fig.~\ref{fig:expl}} presents an explainability analysis for clinical features. The SHapley Additive exPlanations (SHAP)~\cite{lundberg2017unified} values were computed to rank the 20 most significant clinical features contributing to the multimodal model's predictions. Among these, the number of induction chemotherapy cycles emerged as the most influential feature, emphasizing the strong impact of this therapy~\cite{eberhardt1998preoperative} on treatment outcomes. The diagnosis type also played a crucial role, with squamous cell carcinoma appearing to positively influence the pR predictions~\cite{qu2019pathologic}, in contrast to adenocarcinoma. Another highly relevant factor was the number of days of treatment interruption, which underlines the importance of therapy continuity in achieving positive outcomes. The radiotherapy technique also influenced the predictions, underscoring the relevance of VMAT technique~\cite{otto2008volumetric}. Notably, a high number of cigarettes smoked per day emerged as an influential variable for positive pR prediction, reflecting the high smoking prevalence in our dataset among patients with squamous cell carcinoma (positively associated with pR).
\subsection{\textit{Fusion Strategy}}
\begin{figure*}[t] 
    \centering   
    \includegraphics[width=0.94\textwidth]{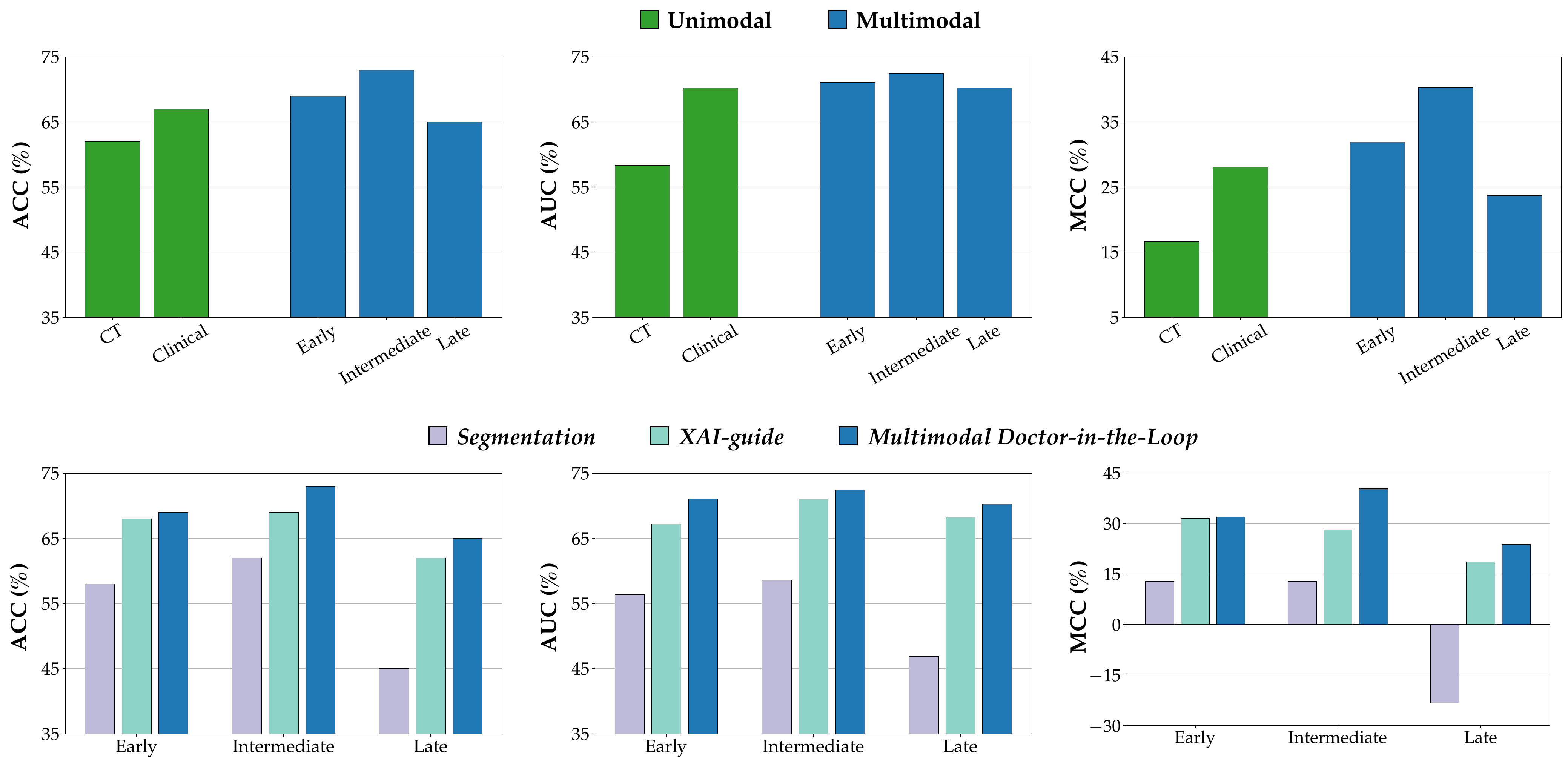} 
    \caption{{Top row: Comparison between the unimodal models (CT and clinical) and the multimodal models trained via early, late and intermediate fusion, with intermediate fusion representing the proposed \textit{Multimodal Doctor-in-the-Loop}. Bottom row: Comparison between \textit{Multimodal Doctor-in-the-Loop}, \textit{Segmentation} and \textit{XAI-guide} approaches via early, late and intermediate fusion.}} 
    \label{fig:plot1} 
\end{figure*}

\hyperref[table:performance_results] {Table~\ref{table:performance_results}} and the top row of \hyperref[fig:plot1] {Fig.~\ref{fig:plot1}} illustrate the impact of early, intermediate, and late fusion strategies on model performance, when integrating the unimodal \textit{Doctor-in-the-Loop} and clinical models. Among the fusion strategies, intermediate fusion consistently achieves the best results, demonstrating its effectiveness in integrating multimodal information. Specifically, the proposed \textit{Multimodal Doctor-in-the-Loop}, which employs intermediate fusion, outperforms both its early and late fusion versions. 

The higher performance of the unimodal clinical model over the unimodal CT model highlights the critical role of clinical features in predicting pR. However, simply combining clinical with imaging data at the input (early fusion) or decision (late fusion) levels does not fully exploit their potential. Early fusion, where imaging and clinical data are merged at the input level, achieves a moderate improvement over the unimodal models. However, this approach does not fully leverage the complementary nature of the two modalities, as it forces them into a shared representation at the data level, potentially losing modality-specific information. On the other hand, late fusion, which combines predictions from separate unimodal models, performs the worst among the fusion strategies. This suggests that treating the two modalities independently until the final decision limits the ability to capture meaningful cross-modal interactions. Intermediate fusion achieves an optimal balance by preserving modality-specific processing initially and enabling joint learning at later stages. This allows for deeper interaction between clinical and imaging features, effectively capturing complex multimodal relationships, thus enhancing predictive performance and model robustness.

\subsection{\textit{Multimodal Doctor-in-the-Loop vs Ablation Studies}}
To further assess the impact of the \textit{Multimodal Doctor-in-the-Loop} method, we compare it against two ablation strategy approaches: one leveraging only explainability guidance (\textit{XAI-guide}) and another relying solely on expert-segmented lesion inputs (\textit{Segmentation}). 
This comparison is performed across all fusion strategies. \hyperref[table:performance_results]{Table~\ref{table:performance_results}} and the bottom row of \hyperref[fig:plot1]{Fig.~\ref{fig:plot1}} show that, across all fusion approaches, the \textit{Doctor-in-the-Loop} approach consistently achieves superior performance. Among the fusion strategies, intermediate fusion stands out, demonstrating its effectiveness in integrating multimodal information. 
Comparing \textit{Multimodal Doctor-in-the-Loop} with \textit{XAI-guide} highlights a key insight: while segmentation guidance improves explainability, relying only on it, without the benefit of GL, limits the model's ability to fully exploit the available data, ultimately leading to lower performance.
Similarly, the comparison with \textit{Segmentation} highlights that relying solely on expert-provided segmentation inputs, without XAI and GL, is not sufficient to achieve optimal performance. This emphasizes the importance of \textit{Doctor-in-the-Loop} methodology, which incorporates insights from the global image while progressively directing the model's focus toward lung and lesion views.

\section{Conclusion}
This study highlights the crucial role of integrating multimodal data and selecting an effective fusion strategy in multimodal learning for predicting pR in NSCLC patients undergoing NAT. Intermediate fusion emerged as the optimal strategy, maintaining the strengths of modality-specific processing while facilitating meaningful cross-modal interactions, resulting in a more accurate and reliable prediction model. Furthermore, integrating \textit{intrinsic} explainability and clinician domain knowledge through our proposed \textit{Multimodal Doctor-in-the-Loop} method significantly enhanced the clinical relevance and explainability of the model.
Future work should explore further validation of our method on larger, multicentric datasets to improve generalizability. Additional research could investigate the integration of longitudinal patient data to enhance predictive accuracy and robustness. Finally, expanding our approach to other clinical endpoints or cancer subtypes could extend its applicability in personalized oncology.
\section*{Acknowledgment}
Alice Natalina Caragliano is a Ph.D. student enrolled in the National Ph.D. in Artificial Intelligence, XXXIX cycle, course on Health and life sciences, organized by Università Campus Bio-Medico di Roma.
This work was partially founded by: 
i) PNRR 2022 MUR P2022P3CXJ-PICTURE; 
ii) PNRR MUR project PE0000013-FAIR;
iii) PNRR M6/C2 project PNRR-MCNT2-2023-123777;
iv) Cancerforskningsfonden Norrland project MP23-1122;
v) Kempe Foundation project JCSMK24-0094.
Resources are provided by the National Academic Infrastructure for Supercomputing in Sweden (NAISS) and the Swedish National Infrastructure for Computing (SNIC) at Alvis @ C3SE. 

\bibliographystyle{IEEEtran}
\bibliography{bib_multi_ijcnn.bib}

\begin{thebibliography}{10}
\providecommand{\url}[1]{#1}
\csname url@samestyle\endcsname
\providecommand{\newblock}{\relax}
\providecommand{\bibinfo}[2]{#2}
\providecommand{\BIBentrySTDinterwordspacing}{\spaceskip=0pt\relax}
\providecommand{\BIBentryALTinterwordstretchfactor}{4}
\providecommand{\BIBentryALTinterwordspacing}{\spaceskip=\fontdimen2\font plus
\BIBentryALTinterwordstretchfactor\fontdimen3\font minus \fontdimen4\font\relax}
\providecommand{\BIBforeignlanguage}[2]{{%
\expandafter\ifx\csname l@#1\endcsname\relax
\typeout{** WARNING: IEEEtran.bst: No hyphenation pattern has been}%
\typeout{** loaded for the language `#1'. Using the pattern for}%
\typeout{** the default language instead.}%
\else
\language=\csname l@#1\endcsname
\fi
#2}}
\providecommand{\BIBdecl}{\relax}
\BIBdecl

\bibitem{cancernet2022}
A.~C. Society, ``Key statistics for lung cancer,'' available online at: \url{https://www.cancer.org/cancer/types/lung-cancer/about/key-statistics.html}.

\bibitem{betticher2006prognostic}
D.~Betticher \emph{et~al.}, ``{Prognostic factors affecting long-term outcomes in patients with resected stage IIIA pN2 non-small-cell lung cancer: 5-year follow-up of a phase II study},'' \emph{British journal of cancer}, vol.~94, 2006.

\bibitem{rosner2022association}
S.~Rosner \emph{et~al.}, ``{Association of pathologic complete response and long-term survival outcomes among patients treated with neoadjuvant chemotherapy or chemoradiotherapy for NSCLC: a meta-analysis},'' \emph{JTO Clinical and Research Reports}, vol.~3, no.~9, p. 100384, 2022.

\bibitem{hellmann2014pathological}
M.~D. Hellmann \emph{et~al.}, ``{Pathological response after neoadjuvant chemotherapy in resectable non-small-cell lung cancers: proposal for the use of major pathological response as a surrogate endpoint},'' \emph{The lancet oncology}, vol.~15, no.~1, pp. e42--e50, 2014.

\bibitem{provencio2020neoadjuvant}
M.~Provencio \emph{et~al.}, ``{Neoadjuvant chemotherapy and nivolumab in resectable non-small-cell lung cancer (NADIM): an open-label, multicentre, single-arm, phase 2 trial},'' \emph{The Lancet Oncology}, vol.~21, 2020.

\bibitem{weissferdt2020agreement}
A.~Weissferdt \emph{et~al.}, ``{Agreement on major pathological response in NSCLC patients receiving neoadjuvant chemotherapy},'' \emph{Clinical lung cancer}, vol.~21, no.~4, pp. 341--348, 2020.

\bibitem{coroller2017radiomic}
T.~P. Coroller \emph{et~al.}, ``{Radiomic-based pathological response prediction from primary tumors and lymph nodes in NSCLC},'' \emph{Journal of Thoracic Oncology}, vol.~12, no.~3, pp. 467--476, 2017.

\bibitem{liu2023development}
C.~Liu \emph{et~al.}, ``{Development and validation of a radiomics-based nomogram for predicting a major pathological response to neoadjuvant immunochemotherapy for patients with potentially resectable non-small cell lung cancer},'' \emph{Frontiers in Immunology}, vol.~14, p. 1115291, 2023.

\bibitem{sun2024pet}
Y.~Sun \emph{et~al.}, ``{PET/CT radiomics and deep learning in the diagnosis of benign and malignant pulmonary nodules: progress and challenges},'' \emph{Frontiers in Oncology}, vol.~14, p. 1491762, 2024.

\bibitem{cellina2022artificial}
M.~Cellina \emph{et~al.}, ``{Artificial intelligence in lung cancer imaging: unfolding the future},'' \emph{Diagnostics}, vol.~12, no.~11, p. 2644, 2022.

\bibitem{qu2024non}
W.~Qu \emph{et~al.}, ``{Non-invasive prediction for pathologic complete response to neoadjuvant chemoimmunotherapy in lung cancer using CT-based deep learning: a multicenter study},'' \emph{Frontiers in Immunology}, vol.~15, p. 1327779, 2024.

\bibitem{ye2024non}
G.~Ye \emph{et~al.}, ``{Non-invasive multimodal CT deep learning biomarker to predict pathological complete response of non-small cell lung cancer following neoadjuvant immunochemotherapy: a multicenter study},'' \emph{Journal for Immunotherapy of Cancer}, vol.~12, no.~9, 2024.

\bibitem{sangeetha2024enhanced}
S.~Sangeetha \emph{et~al.}, ``{An enhanced multimodal fusion deep learning neural network for lung cancer classification},'' \emph{Systems and Soft Computing}, vol.~6, p. 200068, 2024.

\bibitem{guarrasi2025systematic}
V.~Guarrasi \emph{et~al.}, ``{A systematic review of intermediate fusion in multimodal deep learning for biomedical applications},'' \emph{Image and Vision Computing}, p. 105509, 2025.

\bibitem{francesconi2025class}
A.~Francesconi \emph{et~al.}, ``{Class balancing diversity multimodal ensemble for Alzheimer’s disease diagnosis and early detection},'' \emph{Computerized Medical Imaging and Graphics}, p. 102529, 2025.

\bibitem{ruffini2024multi}
F.~Ruffini \emph{et~al.}, ``{Multi-Dataset Multi-Task Learning for COVID-19 Prognosis},'' in \emph{International Conference on Medical Image Computing and Computer-Assisted Intervention}.\hskip 1em plus 0.5em minus 0.4em\relax Springer, 2024, pp. 251--261.

\bibitem{di2025graph}
G.~Di~Teodoro \emph{et~al.}, ``{A graph neural network-based model with Out-of-Distribution Robustness for enhancing Antiretroviral Therapy Outcome Prediction for HIV-1},'' \emph{Comp Medical Imaging Graphics}, vol. 120, 2025.

\bibitem{guarrasi2023multi}
V.~Guarrasi \emph{et~al.}, ``{Multi-objective optimization determines when, which and how to fuse deep networks: An application to predict COVID-19 outcomes},'' \emph{Computers in Biology and Medicine}, vol. 154, 2023.

\bibitem{mogensen2025optimized}
K.~Mogensen \emph{et~al.}, ``{An optimized ensemble search approach for classification of higher-level gait disorder using brain magnetic resonance images},'' \emph{Computers in Biology and Medicine}, vol. 184, p. 109457, 2025.

\bibitem{guarrasi2022optimized}
V.~Guarrasi \emph{et~al.}, ``{Optimized fusion of CNNs to diagnose pulmonary diseases on chest X-Rays},'' in \emph{International Conference on Image Analysis and Processing}.\hskip 1em plus 0.5em minus 0.4em\relax Springer, 2022, pp. 197--209.

\bibitem{caruso2022multimodal}
C.~M. Caruso \emph{et~al.}, ``{A multimodal ensemble driven by multiobjective optimisation to predict overall survival in non-small-cell lung cancer},'' \emph{Journal of Imaging}, vol.~8, no.~11, p. 298, 2022.

\bibitem{lin2022ct}
Q.~Lin \emph{et~al.}, ``{CT-based radiomics in predicting pathological response in non-small cell lung cancer patients receiving neoadjuvant immunotherapy},'' \emph{Frontiers in Oncology}, vol.~12, p. 937277, 2022.

\bibitem{she2022deep}
Y.~She \emph{et~al.}, ``{Deep learning for predicting major pathological response to neoadjuvant chemoimmunotherapy in non-small cell lung cancer: A multicentre study},'' \emph{EBioMedicine}, vol.~86, 2022.

\bibitem{hou2024self}
J.~Hou \emph{et~al.}, ``{Self-eXplainable AI for Medical Image Analysis: A Survey and New Outlooks},'' \emph{arXiv preprint arXiv:2410.02331}, 2024.

\bibitem{guarrasi2024multimodal}
V.~Guarrasi \emph{et~al.}, ``{Multimodal explainability via latent shift applied to COVID-19 stratification},'' \emph{Pattern Recognition}, vol. 156, 2024.

\bibitem{caragliano2025doctor}
A.~N. Caragliano \emph{et~al.}, ``{Doctor-in-the-Loop: An Explainable, Multi-View Deep Learning Framework for Predicting Pathological Response in Non-Small Cell Lung Cancer},'' \emph{arXiv preprint arXiv:2502.17503}, 2025.

\bibitem{bib:selvaraju2017grad}
R.~R. Selvaraju \emph{et~al.}, ``{Grad-cam: Visual explanations from deep networks via gradient-based localization},'' in \emph{Proceedings of the IEEE international conference on computer vision}, 2017, pp. 618--626.

\bibitem{edition2017ajcc}
S.~Edition, S.~Edge, D.~Byrd \emph{et~al.}, ``{AJCC cancer staging manual},'' \emph{AJCC cancer staging manual}, 2017.

\bibitem{zhu2020ct}
Y.~Zhu \emph{et~al.}, ``{A CT-derived deep neural network predicts for programmed death ligand-1 expression status in advanced lung adenocarcinomas},'' \emph{Annals of Translational Medicine}, vol.~8, no.~15, 2020.

\bibitem{lundberg2017unified}
S.~M. Lundberg and S.-I. Lee, ``{A unified approach to interpreting model predictions},'' \emph{Advances Neural Inf Processing Systems}, vol.~30, 2017.

\bibitem{eberhardt1998preoperative}
W.~Eberhardt \emph{et~al.}, ``{Preoperative chemotherapy followed by concurrent chemoradiation therapy based on hyperfractionated accelerated radiotherapy and definitive surgery in locally advanced non-small-cell lung cancer: mature results of a phase II trial.}'' \emph{J Clinical Oncology}, vol.~16, 1998.

\bibitem{qu2019pathologic}
Y.~Qu \emph{et~al.}, ``{Pathologic assessment after neoadjuvant chemotherapy for NSCLC: importance and implications of distinguishing adenocarcinoma from squamous cell carcinoma},'' \emph{Journal of Thoracic Oncology}, vol.~14, no.~3, pp. 482--493, 2019.

\bibitem{otto2008volumetric}
K.~Otto, ``{Volumetric modulated arc therapy: IMRT in a single gantry arc},'' \emph{Medical physics}, vol.~35, no.~1, pp. 310--317, 2008.

\end{thebibliography}
\end{document}